\documentclass[a4paper,fleqn]{cas-dc}

\usepackage[authoryear,longnamesfirst]{natbib}

\usepackage{xcolor}
\usepackage{graphicx}
\usepackage{amsmath,amssymb}
\usepackage{booktabs}
\usepackage{tabularx}
\usepackage{array}
\usepackage{multirow}
\usepackage{algorithm}
\usepackage{algpseudocode}
\usepackage{xurl}
\usepackage{hyperref}
\usepackage{microtype}

\hypersetup{
    colorlinks=true,
    linkcolor=blue,
    citecolor=blue,
    urlcolor=cyan,
    pdfauthor={M. M. Golam Hafiz, A.K.M. Shahriar Emtiaz, Safia Akter Simi, Md. Rakibul Islam, M. F. Mridha}
}

\begin{document}
\let\WriteBookmarks\relax

\shorttitle{Reliable Prerequisite-Aware Learning Path Recommendation}
\shortauthors{M.\,M. Golam Hafiz et al.}

\title[mode=title]{Reliable Prerequisite-Aware Learning Path Recommendation under Uncertain Knowledge Structures: Long-Horizon Evaluation and Cross-Dataset Validation}

\author[1]{M.\,M. Golam Hafiz}[orcid=0009-0003-2795-1033]
\ead{rafihafiz2016@gmail.com}

\author[1]{A.K.M. Shahriar Emtiaz}
\ead{akmshahriaremtiaz@gmail.com}

\author[1]{Safia Akter Simi}
\ead{safiasimi17@gmail.com}

\author[2]{Md. Rakibul Islam}
\cormark[1]
\ead{md.rakibul.islam@mdu.se}

\author[1]{M.\,F. Mridha}[orcid=0000-0001-5738-1631]
\ead{firoz.mridha@aiub.edu}

\affiliation[1]{
    organization={Department of Computer Science,
    American International University-Bangladesh (AIUB)},
    city={Dhaka},
    country={Bangladesh}
}

\affiliation[2]{
    organization={Department of Computer Science and Engineering,
    Mälardalen University},
    city={Västerås},
    state={Västmanland},
    country={Sweden}
}

\cortext[cor1]{Corresponding author: Md. Rakibul Islam
(\href{mailto:md.rakibul.islam@mdu.se}{md.rakibul.islam@mdu.se}).}

\begin{abstract}
Learning-path recommendation often depends on knowledge and prerequisite structures whose reliability is uncertain, while high accuracy in predicting the next resource does not necessarily establish reliable long-horizon path behaviour. This study develops a reliability-oriented prerequisite-aware framework that jointly examines immediate next-resource ranking, uncertain prerequisite structures, hard, soft and hybrid constraint semantics, graph robustness, and autoregressive path generation under course-level confirmatory evaluation and independent cross-dataset validation. Development used 3,187 MOOCCubeX courses, followed by a frozen confirmatory evaluation on 541 held-out courses and 37,747 transitions. The primary MOOCCubeX task is structural and course-prefix-conditioned rather than learner-personalised. Junyi provides independent structure-/hierarchy-aware external validation, uses hierarchy metadata rather than an explicit prerequisite graph and is not a zero-shot transfer setting. On the frozen MOOCCubeX evaluation, XGBoost pairwise was the strongest observed immediate ranker, with mean reciprocal rank (MRR) of 0.9048, whereas the proposed prerequisite-aware graph model achieved 0.7707. In the controlled graph-variant development ablation, the no-graph condition produced the strongest immediate ranking result, and immediate prediction showed strong dependence on permitted title/path-similarity information. Autoregressive evaluation exposed path properties not captured by one-step ranking. On Junyi, Markov achieved the strongest observed MRR of 0.4453, compared with 0.3954 for the proposed graph model, with metric-specific differences across ranking measures. Overall, immediate predictive accuracy, prerequisite-structure reliability, and long-horizon path quality should be treated as complementary rather than interchangeable evaluation dimensions in educational recommendation.
\end{abstract}

\maketitle
\hypersetup{pdfauthor={M. M. Golam Hafiz, A.K.M. Shahriar Emtiaz, Safia Akter Simi, Md. Rakibul Islam, M. F. Mridha}}

\section{Introduction}

Digital learning environments expose learners to large and heterogeneous collections of courses, concepts and resources, but access alone does not determine a coherent route through them. Learning-path recommendation addresses this problem by organising learning objects into ordered sequences intended to support progress towards learning goals \citep{Nabizadeh2020LearningPathSurvey}. This sequence-level view differs from isolated item recommendation because the usefulness of a resource depends partly on where it appears within a broader learning process. Early full-path models accordingly treated recommendation as the prediction and selection of complete learning-resource sequences rather than only the next item \citep{Zhou2018FullPathLSTM}. In education, ordering is consequential because concepts and learning objects can exhibit dependencies; a relevant resource may still be poorly placed if required prior knowledge has not been encountered.

Learning-path construction is therefore not only a relevance-ranking problem but also a constrained structural decision problem. Knowledge-map approaches have shown that path selection may need to balance starting and target knowledge units, learning time, prior exposure and scenario-specific constraints \citep{Zhu2018MultiConstraintKnowledgeMap}. Knowledge-graph formulations represent learning objects and their semantic or dependency relations explicitly, enabling paths to be generated over richer structural representations \citep{Shi2020MultidimensionalKG}. These representations make prerequisite-aware progression computationally expressible, but a constructed graph edge is not necessarily pedagogically authoritative, because behavioural or statistical associations may not represent valid prerequisite relations \citep{Zeng2026ConceptualPathBias}. The reliability of the structure used to constrain recommendations is therefore part of the recommendation problem, not merely a preprocessing concern.

Recent work has expanded this structural perspective in several directions. Knowledge-aware sequence models combine course graphs with sequential representations to capture course information and dependencies beyond adjacent interactions \citep{Deng2023KnowledgeAwareCourseRec}, while context-aware path models use multidimensional knowledge graphs to represent richer educational relationships \citep{Zhang2023FineGrainedKGPath}. Reinforcement-learning studies formulate path generation as sequential decision-making, including methods that address extrapolation error in offline educational recommendation \citep{Yun2024DoublyConstrainedOfflineRL} and methods that balance local and longer-term objectives through multi-behaviour state modelling \citep{Ma2024MultiBehaviorPathRec}. Other approaches cast complete-path construction as multigranularity constrained optimisation \citep{Zheng2025MultigranularityLPR}, or combine knowledge graphs, forgetting factors and reinforcement learning to coordinate new and previously learned concepts \citep{Fan2026ForgettingKG}. These developments increase modelling capacity, but also increase dependence on the quality and interpretation of structural signals.

That dependence creates a reliability problem. Confidence-aware knowledge-graph representation shows that entity and relation information can be heterogeneous and uneven in reliability, motivating confidence-sensitive aggregation \citep{Huang2024KGConfidenceEmbedding}. More directly, recent learning-path research on conceptual path bias argues that behavioural transitions or statistical associations should not automatically be interpreted as pedagogically valid prerequisites; structural errors can propagate into downstream path decisions \citep{Zeng2026ConceptualPathBias}. Immediate next-resource accuracy therefore cannot by itself establish the structural validity of a recursively generated path \citep{Zeng2026ConceptualPathBias}. Despite substantial progress, evidence remains limited on how uncertainty in prerequisite structure and alternative constraint semantics jointly affect immediate ranking and long-horizon path behaviour under leakage-controlled held-out-course, perturbation and cross-dataset evaluation.

The present study develops and evaluates a reliability-oriented prerequisite-aware framework combining prerequisite-graph auditing and refinement, alternative graph variants and uncertainty-aware structural evaluation, hard, soft and hybrid prerequisite policies, resource ranking, autoregressive path generation, controlled graph perturbation, course-level held-out evaluation and independent external validation. The objective is not to maximise one ranking metric, but to examine whether structurally informed recommendation remains reliable under imperfect prerequisite knowledge, whether immediate ranking and long-horizon path behaviour support the same conclusions, and whether observed behaviours persist beyond the primary MOOCCubeX setting. The study addresses three research questions:

\noindent\textbf{RQ1.} How do prerequisite-graph quality and hard, soft or hybrid constraint semantics affect next-resource ranking and generated learning paths?

\noindent\textbf{RQ2.} What relationship exists between immediate next-resource accuracy and long-horizon prerequisite-aware path quality, including structural validity, coverage, redundancy, diversity and progression?

\noindent\textbf{RQ3.} How robust and generalisable are the observed behaviours across controlled prerequisite-graph perturbations, held-out courses and the independent Junyi structure-/hierarchy-aware external-validation setting?

This paper makes the following contributions:

\begin{itemize}

    \item A reliability-oriented formulation that separates immediate next-resource relevance from long-horizon structural path quality and compares alternative prerequisite-constraint semantics rather than assuming a single authoritative hard graph.

    \item A prerequisite-structure audit and robustness framework covering graph refinement, edge uncertainty, graph variants and controlled missing, noisy and reversed structural perturbations.

    \item An autoregressive learning-path evaluation protocol extending beyond one-step ranking to examine prerequisite violations, coverage, redundancy, diversity, structural depth, efficiency and agreement with authored continuation patterns.

    \item A leakage-controlled evaluation combining repeated course-level development validation, uncertainty estimates and paired statistics, strong conventional and neural baselines, independent Junyi structure-/hierarchy-aware validation, and a pre-frozen MOOCCubeX confirmatory evaluation while retaining mixed findings where strong conventional or neural ranking baselines outperform the proposed graph model in immediate prediction.

\end{itemize}

The remainder of this paper is organised as follows. Section 2 reviews related work. Section 3 formulates the problem, presents the prerequisite-aware framework, and describes the experimental protocol. Section 4 reports the results. Section 5 discusses their educational and methodological implications, including limitations and future work, and Section 6 concludes the study.

\section{Related Work}

\subsection{Knowledge-Structured and Constraint-Aware Learning-Path Recommendation}

Learning-path recommendation has long been formulated at the sequence level rather than solely as isolated item selection. Early full-path approaches used learner histories and sequence models to predict complete routes through learning resources; for example, clustering combined with LSTM modelling was used to predict and select full learning paths for individual learners \citep{Zhou2018FullPathLSTM}. A related adaptive strand incorporated learner-specific constraints into explicit course-graph search. \citet{Nabizadeh2020AdaptiveAuxiliary} used a two-layer course graph to construct paths according to learners' knowledge backgrounds and available time, with auxiliary learning objects introduced when expected progress was not achieved. These approaches established that educational recommendation can target an ordered learning trajectory rather than only the next resource.

Structural representations subsequently made dependencies and path requirements more explicit. Knowledge-map methods formulated path construction under multiple learning scenarios and constraints involving resource organisation, learning behaviour and limited time \citep{Zhu2018MultiConstraintKnowledgeMap}. Knowledge-graph approaches extended this idea by representing several semantic relationships among learning objects and generating paths according to target learning objects and learning needs \citep{Shi2020MultidimensionalKG}. Such work demonstrates that learning-path recommendation is inherently constrained: candidate resources cannot be treated as interchangeable items when their position in a sequence depends on knowledge relations, learner requirements or other path-level conditions.

More recent research has broadened the mechanisms through which complete or adaptive paths are constructed. Multigranularity learning-path recommendation models distinguish knowledge-point and learning-object layers and formulate sequencing across different resource granularities as a constrained optimisation problem \citep{Zheng2025MultigranularityLPR}. Offline reinforcement-learning approaches instead model path recommendation as sequential decision making while constraining policy behaviour to address extrapolation errors associated with historical educational data \citep{Yun2024DoublyConstrainedOfflineRL}. Process-type learning paths provide another formulation: process mining is used to represent sequences together with parallel and selection relationships, while deep knowledge tracing and branch recommendation allow the path to respond to changing learner knowledge states \citep{Zhang2024ProcessTypePath}. These formulations therefore span complete-path optimisation, stepwise adaptation and richer process structures rather than a single definition of path generation.

Across these studies, path quality is multi-objective and structurally constrained, involving dependencies, learner or scenario requirements, granularity, adaptation and sequential decision making. In these strands, the knowledge structure or dependency representation is primarily an input to path construction or optimisation; the downstream consequences of uncertainty in the prerequisite relations themselves are less central to the evaluation. Alongside explicit path-optimisation approaches, recent educational recommendation increasingly includes data-driven sequential, graph-based and reinforcement-learning models.

\subsection{Sequential, Graph-Based, and Learner-State Educational Recommendation}

A substantial parallel literature focuses on predicting suitable courses or resources from interaction sequences rather than constructing a complete educational path. Knowledge-aware sequence modelling combines heterogeneous course information with sequential representations; \citet{Deng2023KnowledgeAwareCourseRec}, for example, embed a course knowledge graph and model learning sequences with bidirectional recurrent, convolutional and feed-forward components for online course recommendation. Hypergraph-based course recommendation similarly represents learners as hyperedges while combining long- and short-term course-sequence information \citep{Wang2022HGNN}. KGAN instead aggregates relation-grouped information from a heterogeneous course graph to estimate learner interests under sparse MOOC interactions \citep{Zhang2023KGAN}. These methods demonstrate the value of structural and sequential signals for course ranking, but course recommendation or next-resource prediction is not equivalent to recursively generating and evaluating a complete learning path.

Graph-based educational recommendation has also incorporated richer resource relations and interactive decision making. Multi-knowledge enhanced graph convolution integrates learning-resource--knowledge and learner--knowledge information within a Markov decision formulation for learning-resource recommendation \citep{Dong2024MultiKnowledgeGCN}. HGCR constructs a heterogeneous teacher--course--concept graph, derives representations through graph attention and uses a double-dueling deep Q-network for multi-round interactive course recommendation \citep{Wang2024HGCR}. In learning-path settings, multidimensional knowledge graphs have also been combined with contextual information from learner postings and backgrounds \citep{Zhang2023FineGrainedKGPath}, while multi-behaviour transformers and cascading deep Q-networks model learner characteristics and balance local and longer-term objectives \citep{Ma2024MultiBehaviorPathRec}. FKGRec further combines forgetting factors, knowledge-graph features, cognitive-state modelling and reinforcement learning to sequence new and previously learned concepts \citep{Fan2026ForgettingKG}. A general educational knowledge graph may encode heterogeneous semantic, resource, learner or contextual relations and should therefore not automatically be interpreted as an explicit directional prerequisite graph \citep{Zhang2023FineGrainedKGPath,Zeng2026ConceptualPathBias}.

Learner-state modelling forms a related but distinct technical stream. Knowledge tracing explicitly estimates evolving knowledge state from sequences of learner--system interactions \citep{Shen2024KnowledgeTracingSurvey}, while cognitive-diagnosis models infer learner proficiency from response evidence. Confidence-guided multigraph cognitive diagnosis, for instance, models uncertainty in student responses and aggregates different correct- and incorrect-response relations \citep{Su2024ConfidenceMultigraph}. Such learner-state signals support forms of adaptation that differ from a structural recommendation setting based on authored course prefixes and resource/concept information. The distinction is methodological rather than hierarchical: learner-state uncertainty concerns what is known about the learner, whereas structural uncertainty concerns the reliability of the relations used to organise or constrain recommendations. Modern educational recommenders therefore provide increasingly rich sequential, graph and learner representations, but predictive sophistication alone does not establish whether the structural relations guiding recommendations are pedagogically reliable \citep{Zeng2026ConceptualPathBias}.

\begin{table*}[pos=tbp]
\centering
\caption{Representative recent approaches to knowledge-aware educational recommendation and learning-path generation, highlighting differences in structural assumptions and evaluation scope.}
\label{tab:related_work_comparison}
\scriptsize
\setlength{\tabcolsep}{2.8pt}
\renewcommand{\arraystretch}{1.08}
\begin{tabularx}{\textwidth}{
>{\raggedright\arraybackslash}p{1.7cm}
>{\raggedright\arraybackslash}p{2.0cm}
>{\raggedright\arraybackslash}p{2.0cm}
>{\raggedright\arraybackslash}p{2.3cm}
>{\raggedright\arraybackslash}p{2.4cm}
>{\raggedright\arraybackslash}p{2.3cm}
>{\raggedright\arraybackslash}X}
\toprule
Method / year &
Primary task &
Learner state &
Knowledge structure &
Recommendation / path mechanism &
Evaluation focus &
Relevance / distinction \\
\midrule
\citet{Deng2023KnowledgeAwareCourseRec} &
Online course recommendation &
Learning-sequence representation &
Course knowledge graph with heterogeneous course information &
TransD embeddings with BiLSTM, CNN and MLP &
HR, MRR and NDCG for course recommendation on MOOCCube &
Knowledge-aware sequential ranking; not complete-path evaluation \\
\citet{Yun2024DoublyConstrainedOfflineRL} &
Learning-path recommendation &
Historical learner records and educational state representation &
Prerequisite relations; a prerequisite graph is used in the educational data setting &
Doubly constrained offline deep Q-learning &
Offline RL evaluation on D4RL and educational datasets &
Constrained sequential decision making; emphasis on extrapolation error \\
\citet{Ma2024MultiBehaviorPathRec} &
Learning-path recommendation &
Knowledge background, learning style, settings and preferences &
Knowledge-graph-based multi-behaviour representation &
Multi-behaviour transformer with cascading DQN and two-level reward &
Experiments on two real online-learning datasets &
Learner-adaptive long-/short-term objective modelling \\
\citet{Fan2026ForgettingKG} &
Learning-path recommendation &
Cognitive state and forgetting information &
Knowledge graph with antecedent--successor relationships among concepts &
MemGNN concept prediction followed by reinforcement-learning sequencing &
Three-dataset evaluation and case analysis &
Explicitly combines learner-state dynamics, graph structure and path sequencing \\
\citet{Zeng2026ConceptualPathBias} &
Concept- and exercise-level learning-path recommendation &
Evolving proficiency represented by a bidirectional Mamba-based sequence encoder &
LLM-assisted prerequisite and semantic-similarity graphs with directed structural modelling &
Hierarchical concept planning with dual-pool adaptive exercise selection &
Simulator-based offline evaluation on three real-world datasets &
Directly separates behavioural/statistical relations from pedagogical structural validity \\
\bottomrule
\end{tabularx}
\end{table*}

\subsection{Reliability, Knowledge-Structure Uncertainty, and Evaluation Gap}

The preceding literature makes the reliability of structural information a distinct concern from model capacity. In general recommendation, confidence-aware knowledge-graph embedding has been proposed because entity and relation aggregation can be complex, diverse and imbalanced; confidence mechanisms can therefore alter how graph information contributes to representation learning \citep{Huang2024KGConfidenceEmbedding}. This provides methodological support for treating graph information as unequally reliable, but it does not specifically establish prerequisite-edge validity in educational paths. Educational confidence modelling has likewise appeared in cognitive diagnosis, where response confidence and noise are modelled to improve inference about learner cognition \citep{Su2024ConfidenceMultigraph}. That uncertainty concerns learner evidence, not whether an edge in a prerequisite structure is pedagogically trustworthy.

The structural distinction is sharper in recent learning-path work on conceptual path bias. \citet{Zeng2026ConceptualPathBias} identify the risk that observed learning transitions or statistical associations may be treated as if they were valid prerequisite or similarity relations, and define the downstream path-generation bias caused by incorrect structural relations. This is particularly relevant because an observed \(A\!\rightarrow\!B\) sequence may reflect curriculum order or platform exposure rather than a pedagogical requirement that \(A\) precede \(B\). Consequently, knowledge-structure identification and recommendation cannot be evaluated only through immediate predictive agreement: a structurally erroneous relation can influence later decisions when recommendations are generated recursively.

Related work outside educational learning-path recommendation reinforces the broader methodological distinction between recommendation relevance and path properties. Multi-modal knowledge-aware reinforcement learning uses paths through a knowledge graph to provide explainable recommendation reasoning \citep{Tao2021MKRLN}, while reinforcement-based reasoning-path work evaluates explanation paths using properties such as recency, popularity and diversity in addition to recommendation quality \citep{Balloccu2023ExplanationPathQuality}. These are explanation or reasoning paths, not educational learning paths, and should not be conflated with prerequisite-respecting sequences of learning resources. They nevertheless show that path-level properties can be assessed separately from the relevance of the terminal recommendation.

Prior studies demonstrate substantial progress in full-path recommendation, constrained optimisation, knowledge-aware and graph-based recommendation, learner-state modelling, reinforcement-learning sequencing, confidence-aware representation and path-level reasoning. Less well characterised is how the reliability of the prerequisite structure itself affects both immediate resource ranking and recursively generated learning paths when alternative constraint semantics, controlled structural perturbations, course-level generalisation, statistical uncertainty, strong non-graph baselines and independent external validation are considered jointly. This motivates a reliability-oriented formulation in which predictive relevance, prerequisite feasibility, graph uncertainty and long-horizon path behaviour are treated as complementary rather than interchangeable properties. Accordingly, the framework in Section~3 treats prerequisite structure as an uncertain component of the recommendation problem and evaluates its influence across ranking, constraint enforcement and autoregressive path generation.

\section{Methodology}

\begin{figure*}[pos=tbp]
    \centering
    \includegraphics[
        width=0.90\textwidth,
        keepaspectratio
    ]{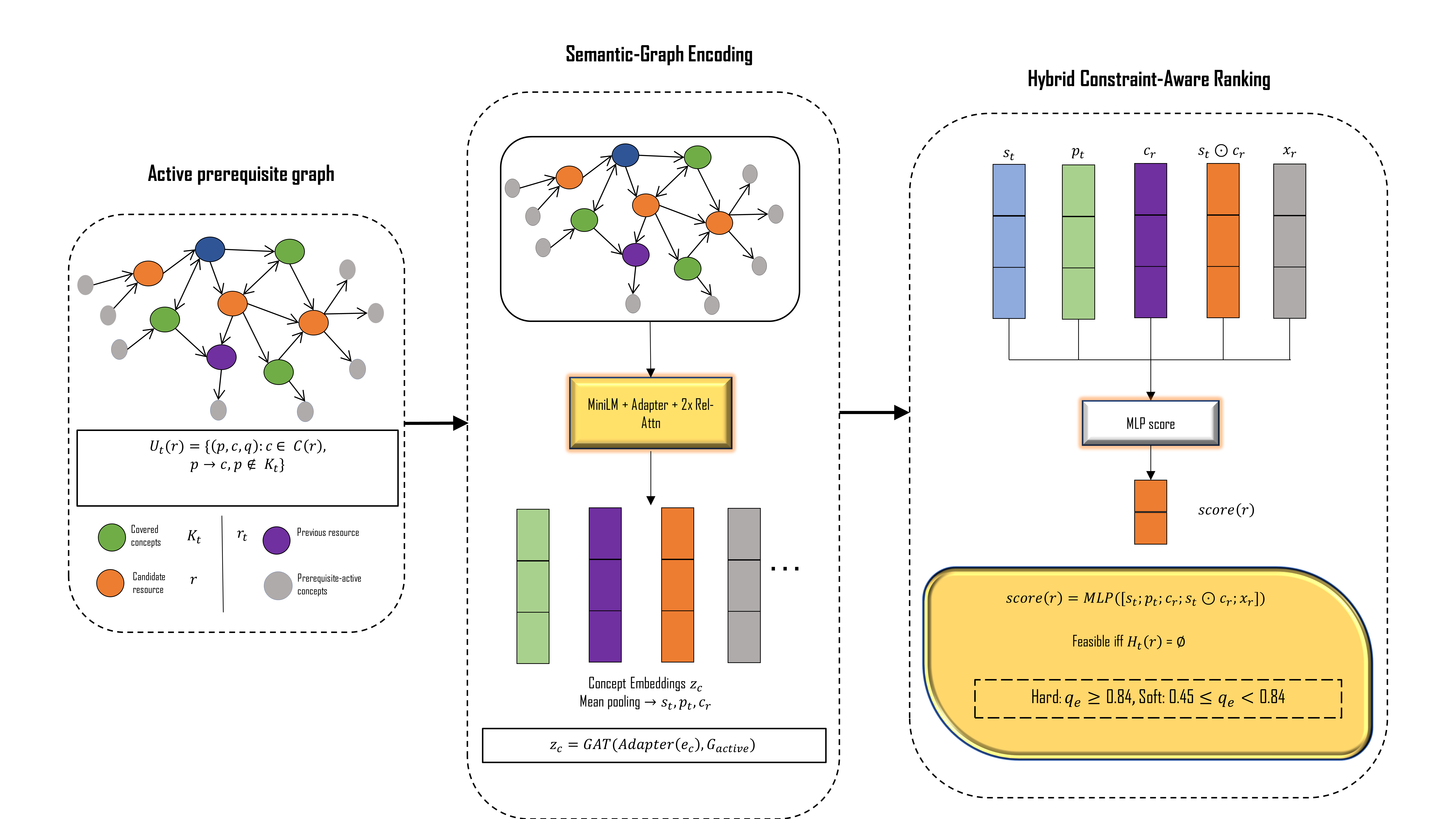}
    \caption{Architecture of the proposed prerequisite-aware ranking model, comprising active prerequisite identification, semantic-graph encoding, and hybrid constraint-aware ranking.}
    \label{fig:method_framework}
\end{figure*}

The methodology comprises a reliability-oriented prerequisite-aware recommendation framework and a protocol for development, robustness analysis, long-horizon evaluation, frozen confirmation, and external validation. Figure~\ref{fig:method_framework} summarises the workflow. The primary MOOCCubeX task is structural, course-prefix-conditioned next-resource ranking rather than learner-personalised recommendation. Prerequisite information is treated as an uncertain structural signal through audited graph variants and hard, soft, and hybrid constraint semantics. Graph-based, neural, classical, and heuristic rankers are developed exclusively on development courses, after which the selected configuration is frozen for the course-disjoint frozen confirmatory evaluation. One-step ranking is also converted into autoregressive learning-path generation. An independent Junyi experiment evaluates the framework under learner sequences and hierarchy-aware structural information. Statistical uncertainty, efficiency measurements, frozen-protocol controls, and execution provenance are incorporated within the same experimental design.

\subsection{Problem Formulation and Study Overview}

For course \(c\), let the authored resource sequence be
\[
R_c=(r_1,r_2,\ldots,r_T).
\]
At recommendation step \(t\), the observed course prefix is
\[
P_t=(r_1,\ldots,r_t),
\]
and the supervised positive target is the authored next resource \(r_{t+1}\). If \(C(r)\) denotes the set of concepts aligned with resource \(r\), the concepts covered by the current prefix are
\[
K_t=\bigcup_{j=1}^{t} C(r_j).
\]
The prerequisite structure is represented by \(G=(V,E)\), where \(V\) is the concept set and each directed edge may carry a confidence value. Given an eligible candidate set \(A_t\), a ranking model assigns
\[
s_{\theta}(P_t,r), \qquad r\in A_t,
\]
before the selected prerequisite policy modifies candidate feasibility or structural penalties.

The supervised transition is therefore \((P_t,r_{t+1})\). Importantly, the positive label represents authored continuation, not a claim that adjacency \(r_t\rightarrow r_{t+1}\) is itself a true pedagogical prerequisite. Authored ordering is used as recommendation supervision and, separately, as supporting evidence during graph auditing; these roles are not conflated.

Evaluation occurs at two levels. At Level~1, each observed transition is treated independently and candidate resources are ranked against the authored next resource. At Level~2, the selected resource is appended to the generated sequence,
\[
P_{t+1}=P_t\oplus\hat r_{t+1},
\qquad
K_{t+1}=K_t\cup C(\hat r_{t+1}),
\]
after which candidates and structural state are recomputed. Repetition produces an autoregressive path \(\hat P\). The central methodological objective is to determine how prerequisite structure and alternative constraint semantics affect these two evaluation levels under uncertain structural information without assuming that stronger one-step prediction necessarily implies stronger long-horizon structural behaviour.

\subsection{MOOCCubeX Corpus, Prerequisite Structure, and Leakage-Controlled Representation}

The main study retains the full eligible MOOCCubeX course/resource corpus rather than globally restricting courses to prerequisite-annotated domains. Courses with at least two authored resources are retained. The prerequisite files for computer science, mathematics, and psychology supply structural annotations only. Development contains 3,187 courses; the course-disjoint frozen confirmatory set contains 541 courses and 37,747 authored next-resource transitions. All model, curriculum, graph-configuration, and hyperparameter selection is restricted to development evidence.

Authored resources are represented as videos, exercises, or other resource types. Video--concept alignment is reconstructed through the concept--video and video-identifier relations; exercise alignment is obtained through the exercise--problem and concept--problem relations. Resource title paths provide content-only textual features. Concept text, formed from available concept names and context, is encoded with \texttt{sentence-\allowbreak{}transformers/\allowbreak{}paraphrase-\allowbreak{}multilingual-\allowbreak{}MiniLM-\allowbreak{}L12-\allowbreak{}v2} using normalised embeddings. Population statistics, including concept popularity, are estimated from fitting courses only. Prerequisite depth is computed from the active soft-or-higher-confidence graph.

Prerequisite names are resolved conservatively within domain-compatible concept sets. Unique exact matches are accepted; non-unique exact matches remain ambiguous. Fuzzy matching requires a score of at least \(92\) with a four-point margin, while semantic recovery requires cosine similarity of at least \(0.82\), a \(0.035\) similarity margin, and a fuzzy floor of \(55\). Duplicate resolved edges are removed. Direct-edge confidence combines mapping confidence, source-prediction evidence, semantic relatedness, and development-course directional evidence:
\[
\begin{aligned}
q_e
&=\operatorname{clip}\bigl(
0.30m_e+0.15+0.20p_e+0.15s_e \\
&\qquad\qquad +0.20d_e,\;0,\;1
\bigr),
\end{aligned}
\]
where \(s_e\) is rescaled semantic similarity and \(d_e=0.5\) when fewer than two courses provide directional evidence. Strong reverse-direction evidence reduces confidence by a factor of \(0.72\). Edges with \(q_e\geq0.84\) are designated hard, those with \(0.45\leq q_e<0.84\) soft, and lower-confidence edges review-only. The lowest-confidence active edge in a detected cycle is softened to review status. Development-derived missing-edge proposals require occurrence in at least two courses, directional support of at least \(0.80\), semantic support of at least \(0.62\), and confidence no greater than \(0.79\); proposals that introduce cycles are rejected from the active graph. These operations are automated and do not constitute expert validation.

The selected ranking representation excludes target-, order-, and course-identity shortcuts. The safe feature vector contains \texttt{type\_\allowbreak{}video}, \texttt{type\_\allowbreak{}exercise}, \texttt{type\_\allowbreak{}other}, \texttt{concept\_\allowbreak{}count}, \texttt{new\_\allowbreak{}concept\_\allowbreak{}gain}, \texttt{prev\_\allowbreak{}jaccard}, \texttt{covered\_\allowbreak{}overlap}, \texttt{train\_\allowbreak{}concept\_\allowbreak{}popularity}, \texttt{depth\_\allowbreak{}norm}, \texttt{depth\_\allowbreak{}smoothness}, \texttt{prereq\_\allowbreak{}penalty}, \texttt{unmet\_\allowbreak{}hard\_\allowbreak{}count}, \texttt{unmet\_\allowbreak{}soft\_\allowbreak{}count}, \texttt{novelty\_\allowbreak{}ratio}, \texttt{prev\_\allowbreak{}title\_\allowbreak{}bigram\_\allowbreak{}jaccard}, and \texttt{candidate\_\allowbreak{}title\_\allowbreak{}length}. The prohibited set is \texttt{after\_\allowbreak{}current}, \texttt{candidate\_\allowbreak{}course\_\allowbreak{}id}, \texttt{candidate\_\allowbreak{}order\_\allowbreak{}idx}, \texttt{order\_\allowbreak{}distance}, \texttt{same\_\allowbreak{}course}, \texttt{target\_\allowbreak{}concepts}, and \texttt{target\_\allowbreak{}overlap}.

\begin{table*}[tbp]
\centering
\caption{Datasets, data splits, candidate construction, knowledge structures, and evaluation protocols used in the study.}
\label{tab:method_datasets}

\footnotesize
\setlength{\tabcolsep}{3.5pt}
\renewcommand{\arraystretch}{1.12}

\begin{tabularx}{\textwidth}{
>{\raggedright\arraybackslash}p{1.85cm}
>{\raggedright\arraybackslash}p{2.05cm}
>{\raggedright\arraybackslash}p{1.85cm}
>{\raggedright\arraybackslash}X
>{\raggedright\arraybackslash}p{2.45cm}
>{\raggedright\arraybackslash}p{2.55cm}
>{\raggedright\arraybackslash}p{2.35cm}}
\toprule
\textbf{Dataset / split} &
\textbf{Study role} &
\textbf{Sequence unit} &
\textbf{Split and evaluation design} &
\textbf{Candidate construction} &
\textbf{Knowledge structure} &
\textbf{Sequential protocol} \\
\midrule

MOOCCubeX development~\cite{Yu2021MOOCCubeX}
&
Model and configuration development
&
Authored course-resource sequence
&
3,187 courses; repeated \(5\times2\) course-level cross-validation; final development-only train/tune fitting
&
Full global catalog; training pool of 256; validation cap of 512; development-evaluation cap of 1,024
&
Original, validated, refined, and no-graph prerequisite variants
&
Development-only rollouts and controlled ablations
\\

MOOCCubeX frozen confirmatory set
&
Frozen confirmatory evaluation
&
Authored course-resource sequence
&
541 disjoint courses; all 37,747 transitions evaluated; no transition sampling
&
Full global catalog; cap of 1,024; development and confirmatory-set catalogs
&
Frozen refined prerequisite graph with prespecified perturbations
&
60 deterministically selected courses; beam search enabled for the first 20; maximum 50 steps
\\

Junyi Academy~\cite{Pojen2020Junyi}
&
Independent external validation
&
Learner exercise sequence
&
36,269 training, 5,255 validation, and 10,151 test learners; 21,083 short histories excluded
&
Full 1,330-item catalog at test time; current exercise excluded
&
\texttt{Info\_Content} hierarchy/tree metadata; not treated as a prerequisite graph
&
1,000 eligible test learners; rollout horizon of 10
\\

\bottomrule
\end{tabularx}
\end{table*}
\subsection{Prerequisite-Aware Graph Transformer and Comparison Models}

The proposed \texttt{ActivePrerequisite\allowbreak{}GraphTransformer} initialises each indexed concept with a fixed normalised MiniLM representation and a trainable low-rank semantic adapter of dimension 64. Graph transformation is restricted to concepts occurring in active prerequisite edges. Two sparse relation-attention layers with four heads and dropout \(0.15\) propagate confidence-weighted forward and reverse graph information; concepts outside the active prerequisite subgraph retain their adapted semantic representation without graph propagation. Concept sets for the covered state, previous resource, and candidate resource are mean-pooled. The ranking head receives the pooled state \(s\), previous-resource representation \(p\), candidate representation \(c\), the interaction \(s\odot c\), and the 16-dimensional safe-feature vector. The frozen main graph model contains 15,911,634 parameters.

The Two-Tower comparator uses the same fixed semantic initialisation and low-rank adapters without prerequisite graph propagation. One tower represents the covered state and previous resource; the second represents the candidate concepts together with safe features. Both outputs are normalised and scored by their dot product.

Classical comparisons comprise deterministic Random-valid, Concept-continuity, Popularity-content, a shortcut-free linear pairwise ranker, and XGBoost pairwise learning-to-rank. Concept-continuity combines previous-resource concept overlap, new-concept gain, depth smoothness, and prerequisite penalty; Popularity-content combines train-derived concept popularity, continuity, and new-concept gain. The linear model uses only the safe features. XGBoost uses the \texttt{rank:pairwise} objective with 350 estimators, maximum depth 7, learning rate \(0.045\), subsampling and column sampling of \(0.85\), and \(L_2\) regularisation of 1.0.

\begin{table*}[pos=tbp]
\centering
\caption{Recommendation models considered in the study and their methodological roles.}
\label{tab:method_models}

\footnotesize
\setlength{\tabcolsep}{4pt}
\renewcommand{\arraystretch}{1.12}

\begin{tabularx}{\textwidth}{
>{\raggedright\arraybackslash}p{2.0cm}
>{\raggedright\arraybackslash}p{2.3cm}
>{\raggedright\arraybackslash}X
>{\raggedright\arraybackslash}X
>{\raggedright\arraybackslash}p{2.4cm}
>{\raggedright\arraybackslash}X}
\toprule
\textbf{Model} &
\textbf{Model family} &
\textbf{Graph usage} &
\textbf{Features / inputs} &
\textbf{Constraint handling} &
\textbf{Comparison purpose} \\
\midrule

Random-valid
&
Heuristic
&
None
&
None
&
Candidate feasibility filter
&
Random feasible baseline
\\

Concept continuity
&
Structural/content heuristic
&
Prerequisite depth and other static graph-derived quantities
&
Selected leakage-safe features
&
Candidate feasibility filter
&
Local concept-continuity baseline
\\

Popularity
&
Statistical/content heuristic
&
None
&
Training-set popularity, continuity, and gain features
&
Candidate feasibility filter
&
Frequency- and content-based baseline
\\

Linear HACR-safe
&
Linear pairwise ranker
&
Leakage-safe graph-derived features
&
Full leakage-safe feature set
&
Candidate feasibility filter
&
Linear leakage-controlled comparator
\\

XGBoost pairwise
&
Tree-based learning-to-rank
&
Leakage-safe graph-derived features
&
Full leakage-safe feature set
&
Candidate feasibility filter
&
Nonlinear pairwise comparator
\\

Two-Tower
&
Neural semantic ranker
&
No graph propagation
&
Semantic representations and leakage-safe features
&
Candidate feasibility filter
&
Neural non-graph comparator
\\

\addlinespace[2pt]
\textbf{Proposed graph model}
&
\textbf{Prerequisite-aware graph neural ranker}
&
\textbf{Refined active prerequisite graph}
&
\textbf{Semantic representations and leakage-safe features}
&
\textbf{Frozen hybrid policy}
&
\textbf{Graph-structural ranking model}
\\

\bottomrule
\end{tabularx}
\end{table*}

\subsection{Constraint Semantics, Candidate Ranking, and Development Training}

For candidate resource \(r\), let
\[
U_t(r)=
\{(p,c,q):c\in C(r),\;p\rightarrow c,\;p\notin K_t\}
\]
denote unmet prerequisite edges. Every aligned candidate concept is checked, including concepts previously encountered elsewhere in the candidate. Under \emph{strict-original}, all unmet edges in the original resolved graph are hard. Under \emph{hard-high-confidence}, edges with \(q\geq0.84\) are hard. Under \emph{soft}, no candidate is excluded and edges with \(q\geq0.45\) contribute a penalty. Under the frozen \emph{hybrid} policy,
\[
\begin{aligned}
H_t(r)&=\{e\in U_t(r):q_e\geq0.84\},\\
S_t(r)&=\{e\in U_t(r):0.45\leq q_e<0.84\}.
\end{aligned}
\]
and
\[
\phi_t(r)=\sum_{e\in H_t(r)\cup S_t(r)}q_e,
\qquad
r\ \text{feasible}\iff H_t(r)=\varnothing.
\]
Hard-infeasible candidates are masked; soft evidence remains available to the scorer through prerequisite-derived safe features.

Development neural training draws candidates from the full-global catalog with a pool cap of 256. The authored positive is retained when eligible, and 12 deterministic structural hard negatives are selected using concept continuity, train-derived popularity, and depth proximity. Repeated-CV validation uses a candidate cap of 512. Checkpoint selection uses deterministic validation subsets of 1,000 transitions with cap 128 during CV; final development fitting uses 3,000 checkpoint transitions with cap 384 and evaluates the complete development tuning set with cap 1024. CV transition caps are 50,000 training and 12,000 validation transitions; final development caps are 120,000 training and 20,000 tuning transitions. The frozen confirmatory evaluation uses cap 1024, the combined development and confirmatory-set resource catalog, and no transition sampling. Already observed prefix resources are excluded; missing and structurally infeasible targets are recorded rather than silently removed.

Neural optimisation uses pairwise logistic loss,
\[
\mathcal{L}_{\mathrm{rank}}
=
\frac{1}{B}\sum_{i=1}^{B}
\operatorname{softplus}\!\left[-(s_i^{+}-s_i^{-})\right].
\]
AdamW uses learning rate \(1.5\times10^{-3}\), weight decay \(10^{-4}\), effective pair batch 4096, mixed precision, and gradient-norm clipping at 5.0. A ReduceLROnPlateau scheduler uses factor 0.5, patience 4, and minimum learning rate \(10^{-5}\). Graph CV runs use at most 48 epochs and Two-Tower CV runs 60 epochs, with 16 optimisation steps per epoch. Final development fitting permits 120 epochs, 24 steps per epoch, validation every three epochs, a minimum of 36 epochs, and patience 20.

Model and curriculum selection uses five course-level folds repeated twice over all 3,187 development courses. The frozen selection is \texttt{graph\_\allowbreak{}no\_\allowbreak{}curriculum}; no information from the confirmatory set participates in this decision. XGBoost training groups contain the positive resource and 24 deterministic negatives and use the same leakage-controlled feature schema.

\begin{algorithm}[t]
\caption{Prerequisite-aware next-resource ranking and autoregressive path generation}
\label{alg:path_generation}
\begin{algorithmic}[1]
\Require Prefix \(P_t\), covered concepts \(K_t\), resource catalog \(\mathcal{R}\), mappings \(C(\cdot)\), graph \(G\), trained scorer \(s_\theta\), policy \(\pi\), candidate cap \(M\), maximum path length \(L\), optional beam parameters
\Ensure One-step ranked list and, when requested, generated path \(\hat P\)
\State Construct the state from \(P_t\), \(K_t\), and the previous resource.
\State Form eligible candidates from \(\mathcal{R}\), excluding previously selected resources and applying the stage-specific cap \(M\).
\For{each candidate \(r\)}
    \State Compute leakage-controlled static features and \(C(r)\).
    \State Obtain adapted semantic concept representations.
    \State Replace prerequisite-active concepts by graph-attention representations.
    \State Compute \(s_\theta(P_t,r)\).
    \State Identify \(H_t(r)\), \(S_t(r)\), and \(\phi_t(r)\) under policy \(\pi\).
    \State Mask hard-infeasible candidates or retain soft penalties as prescribed by \(\pi\).
\EndFor
\State Rank remaining candidates by the resulting scores.
\If{one-step evaluation}
    \State \Return ranked candidate list.
\EndIf
\While{stopping condition is false and path length \(<L\)}
    \State Select the highest-ranked candidate for greedy rollout, or retain the prescribed beam hypotheses.
    \State Append \(\hat r_{t+1}\) and update
    \(P_{t+1}=P_t\oplus\hat r_{t+1}\) and
    \(K_{t+1}=K_t\cup C(\hat r_{t+1})\).
    \State Regenerate candidates and repeat scoring and constraint application.
\EndWhile
\State \Return generated path \(\hat P\).
\end{algorithmic}
\end{algorithm}

\subsection{Graph Reliability, Robustness, and Controlled Ablations}

Graph dependence is examined through four controlled structures. \emph{No-graph} removes all prerequisite edges while retaining the remaining model components. \emph{Original} uses the raw resolved source edges as hard relations without confidence-based repair. \emph{Validated} applies automated mapping, semantic/directional confidence, and cycle softening but adds no missing-edge proposals. \emph{Refined} also incorporates development-derived proposals that satisfy the prespecified support, semantic, and cycle criteria.

Constraint semantics are studied in two complementary ways. First, the same frozen relevance model is evaluated under strict-original, soft, hard-high-confidence, and hybrid inference policies, thereby isolating the constraint layer. Second, each policy is retrained under an equal optimisation budget on the same fixed development fold. Graph-variant retraining is likewise equal-budget. These controlled ablations use 24 epochs and 16 optimisation steps per epoch without outcome-dependent budget differences.

Structural sensitivity is evaluated by perturbing the refined graph. Edge-deletion rates are \(0.10,0.20,0.30\), and \(0.50\); synthetic additions use rates \(0.05,0.10,\) and \(0.20\), confidence 0.65, and soft status; edge reversal uses rates \(0.05,0.10,\) and \(0.20\); and additive Gaussian confidence noise uses standard deviations \(0.10\) and \(0.20\), clipped to \([0,1]\). Each condition is repeated with graph seeds 11, 22, and 33, giving 36 prespecified perturbation conditions. These perturbations modify deployment structure without refitting the frozen main model.

A separate feature-reliance audit retains the same candidates and trained model while masking \texttt{title}, \texttt{popularity}, \texttt{depth}, or prerequisite-static feature groups; an additional concept/graph-only condition masks all 16 safe static features. This audit assesses reliance on permitted information classes without introducing any prohibited target or course-order signal.

\subsection{Autoregressive Learning-Path Generation and Junyi External Validation}

For MOOCCubeX path generation, the initial prefix contains
\[
\left\lceil 0.25\,\lvert R_c\rvert \right\rceil
\]
authored resources while retaining at least one suffix resource. Goal concepts are the union of concepts mapped to the authored suffix. The frozen protocol deterministically selects 60 courses from the frozen confirmatory set; the first 20 are also eligible for beam search. At each step, resources covering currently unmet goal concepts are prioritised and the remaining pool is deterministically filled from the global catalog to a maximum of 1024 candidates. Previously selected resources are prohibited.

Rollout selection uses
\[
\begin{aligned}
J_t(r)
&= s_\theta(P_t,r)+1.5\,g_t(r) \\
&\quad -0.50\,\phi_t(r)-0.25\,\rho_t(r),
\end{aligned}
\]
where \(g_t(r)\) is the fraction of goal concepts newly covered and \(\rho_t(r)\) is candidate-concept redundancy with the current state. Greedy rollout selects the maximum-\(J_t\) candidate; beam search uses width 5 and expands the top 20 candidates per hypothesis. Generation terminates at goal coverage \(0.95\), candidate exhaustion, or 50 steps. The protocol selection of 60 courses is distinct from realised evaluable path coverage, which is reported separately. Retained path measures include path length, goal coverage, termination success, prerequisite-violation rate, redundancy, unique concepts, concept diversity per step, depth progression, mean and maximum prerequisite depth, resource repetition, path efficiency, and authored-suffix resource Jaccard overlap.

Junyi provides an independent structure-/hierarchy-aware validation regime based on real learner sequences. Preprocessing yields 16,217,311 raw problem attempts, 8,149,881 collapsed exercise visits, and 1,330 catalog items. A deterministic stable-hash split contains 36,269 training learners, 5,255 validation learners, 10,151 test learners, and 21,083 learners excluded for insufficient history. Training contains 5,486,792 transitions; deterministic caps retain 80,000 validation and 250,000 test transitions. The latter contain 10,131 unique represented test learners, which is distinct from the 10,151-learner test split. Test ranking uses the complete catalog with only the current exercise excluded and no sampled negatives.

Junyi neural models use a 12-step interaction history, item and correctness representations, and Info\_Content difficulty, learning-stage, and hierarchy metadata. The hierarchy-aware model performs two graph-smoothing operations over the hierarchy affinity structure; the no-graph counterpart retains the same sequential representation without this propagation. Training uses 63 sampled negatives, cross-entropy loss, batch size 4096, AdamW with learning rate \(2\times10^{-3}\) and weight decay \(10^{-5}\), at most eight epochs, and validation patience 2. Ranking comparisons comprise \texttt{proposed\_\allowbreak{}graph}, \texttt{proposed\_\allowbreak{}graph\_\allowbreak{}soft}, \texttt{two\_\allowbreak{}tower\_\allowbreak{}no\_\allowbreak{}graph}, Markov, popularity, hierarchy continuity, and random. The soft variant adds a hierarchy-affinity prior of weight 0.20. For path evaluation, 1,000 eligible test learners are selected deterministically and rolled out for up to 10 steps using all models except random; generated exercises are appended to history with correctness fixed to 1.0 for the subsequent simulated state update. External path measures are path length, exact-step agreement, observed-suffix recall, uniqueness, hierarchy coherence, and branch diversity. The Junyi hierarchy is tree metadata and is not interpreted as an explicit prerequisite graph.

\subsection{Evaluation Metrics, Statistical Analysis, Efficiency, and Reproducibility}

Next-resource evaluation reports HR@1, HR@5, HR@10, MRR, NDCG@10, top-one prerequisite violation rate (Viol@1), average candidate count, infeasible-target rate, and missing-target rate. With \(N\) transitions and target rank \(k_i\),
\[
\mathrm{MRR}=\frac{1}{N}\sum_{i=1}^{N}\frac{\mathbb{I}[k_i<\infty]}{k_i}.
\]
HR@\(\!K\) records whether \(k_i\leq K\); NDCG@10 uses the single relevant authored target. Viol@1 is the fraction of top-ranked resources that fail the active policy's hard-feasibility test. The primary MOOCCubeX analysis retains all transitions, including structurally infeasible or missing targets with zero retrieval credit where appropriate. Feasible-only, prerequisite-graph-covered, and concept-mapped subsets are retained as companion analyses rather than replacements for the all-transition evaluation.

Development uncertainty is estimated by course-cluster bootstrap with 2,000 repetitions over repeated course-level validation. Paired comparisons are formed from per-course mean reciprocal-rank differences on common courses; where specified, two-sided Wilcoxon signed-rank and sign tests are used with Holm adjustment. The frozen confirmatory evaluation includes all 37,747 transitions and again treats course as the resampling and paired-comparison unit, with 2,000 bootstrap repetitions. Junyi uncertainty is computed at learner level using 2,000-repetition learner bootstraps and paired learner-level bootstrap comparisons.

Efficiency evidence records training and inference wall-clock time, CPU resident-set memory where measured, GPU allocated memory, parameter count, and candidate throughput. Approximate GPU energy is obtained by trapezoidal integration of \texttt{nvidia-smi} power draw and excludes CPU and wider system energy. The main graph model contains 15,911,634 parameters. Recommendation explanation artifacts record satisfied and unmet prerequisites, new concepts, continuity, redundancy, prerequisite depth, resource type/title evidence, and rollout-objective components. Generated review templates are system artifacts only and are not treated as completed expert, instructor, or trust validation.

The MOOCCubeX implementation uses deterministic seed 42, PyTorch mixed-precision neural training in a dual-Tesla-T4 environment, CPU XGBoost, saved best/last model states, resumable checkpoints, deterministic course splits and candidate construction, protocol manifests, and SHA-256 identity records. The selected development configuration and model states were frozen before confirmatory scoring; the confirmatory protocol fixed full-global ranking, candidate cap 1024, no transition sampling, and 2,000 bootstrap repetitions, with no post-holdout model or configuration selection. Junyi uses its independent deterministic seed \(20260923\) and separately trained external models.

For transparency, Notebook~03 is a recovery execution of the already-opened confirmatory holdout. Notebook03-v1 crossed the one-way holdout gate but failed before ranking or path scoring. The recovery retained the same pre-frozen Notebook~01B model/configuration, the same deterministically reconstructed confirmatory split and transitions, and the original frozen protocol. The failed v1 output was not used as an input, and no retraining, tuning, model selection, protocol change, confirmatory-split change, or post-holdout selection was performed.

\section{Results}

Results are reported in the order of development validation and model freezing, frozen confirmatory next-resource ranking, structural and robustness analyses, autoregressive path evaluation, and Junyi structure-/hierarchy-aware external validation with secondary efficiency and explanation evidence. Development evidence was used to freeze the graph configuration before confirmatory evaluation. All confirmatory outcomes are reported as observed under the frozen protocol and were not used for post-holdout model or configuration selection.

\subsection{Development Validation and Frozen Model Selection}

Development comprised 3,187 courses evaluated through five course-level folds repeated twice. Among the graph curriculum configurations, \texttt{graph\_\allowbreak{}no\_\allowbreak{}curriculum} obtained a transition-level MRR of 0.8128 and a course-cluster mean MRR of 0.7919 (95\% CI: 0.7849--0.7982), compared with 0.7920 and 0.7780 (0.7704--0.7849), respectively, for \texttt{graph\_\allowbreak{}adaptive\_\allowbreak{}balanced}, and 0.7664 and 0.7539 (0.7461--0.7609) for \texttt{graph\_\allowbreak{}static\_\allowbreak{}balanced}. Under the pre-specified development selection rule, the no-curriculum graph configuration therefore provided the strongest graph-model development criterion and was frozen for subsequent evaluation.

Course-level paired comparisons were consistent with this ordering among the graph curriculum variants. Relative to the adaptive-balanced graph configuration, \texttt{graph\_\allowbreak{}no\_\allowbreak{}curriculum} had a mean per-course MRR difference of 0.0139, with 1,686 wins, 1,343 losses and 158 ties across 3,187 courses (Wilcoxon \(p=1.30\times10^{-12}\)). The difference from the static-balanced configuration was 0.0379, with 1,981 wins, 1,037 losses and 169 ties (\(p=2.21\times10^{-101}\)).

The broader development comparison did not make the frozen graph model the highest-ranking predictor. Course-cluster MRR was 0.8905 for XGBoost pairwise, 0.8011 for Two-Tower, 0.7968 for Linear HACR-safe and 0.7919 for the selected graph configuration. These comparator results were retained as part of the controlled evaluation rather than used to replace the pre-specified graph-model selection procedure. Courses in the confirmatory set remained excluded from all development selection decisions.

\subsection{Frozen Confirmatory Next-Resource Ranking}

The confirmatory evaluation contained 541 held-out MOOCCubeX courses and all 37,747 held-out transitions. Table~\ref{tab:confirmatory_ranking} reports the principal all-transition ranking metrics together with course-cluster MRR uncertainty. XGBoost pairwise achieved the highest observed immediate next-resource ranking performance among the compared models, with MRR 0.9048, HR@1 0.8843 and NDCG@10 0.9114. Linear HACR-safe and Two-Tower obtained MRR values of 0.8259 and 0.8243, respectively. The proposed graph model obtained MRR 0.7707, HR@1 0.7269 and NDCG@10 0.7853. Concept-continuity, Random-valid and Popularity-content yielded substantially lower immediate-ranking values.

\begin{table*}[tbp]
\centering
\caption{Frozen confirmatory next-resource ranking performance on 541 held-out MOOCCubeX courses (37,747 transitions). Transition-level ranking metrics and course-cluster MRR are reported separately.}
\label{tab:confirmatory_ranking}

\footnotesize
\setlength{\tabcolsep}{3pt}
\renewcommand{\arraystretch}{1.12}

\begin{tabular*}{\textwidth}{
@{\extracolsep{\fill}}
l
c c c c c c
c
@{}
}
\toprule
\textbf{Model} &
\textbf{HR@1} &
\textbf{HR@5} &
\textbf{HR@10} &
\textbf{MRR} &
\textbf{NDCG@10} &
\textbf{Viol@1} &
\shortstack{\textbf{Course-cluster MRR}\\\textbf{[95\% CI]}} \\
\midrule

Random-valid
& 0.0012
& 0.0048
& 0.0099
& 0.0076
& 0.0047
& 0.0000
& 0.0072 [0.0068, 0.0077] \\

Popularity-content
& 0.0001
& 0.0004
& 0.0007
& 0.0044
& 0.0004
& 0.0000
& 0.0039 [0.0035, 0.0044] \\

Concept-continuity
& 0.0456
& 0.0565
& 0.0588
& 0.0530
& 0.0523
& 0.0000
& 0.0442 [0.0353, 0.0538] \\

Linear HACR-safe
& 0.8069
& 0.8448
& 0.8559
& 0.8259
& 0.8318
& 0.0000
& 0.7985 [0.7831, 0.8140] \\

Two-Tower
& 0.7996
& 0.8498
& 0.8616
& 0.8243
& 0.8321
& 0.0000
& 0.8023 [0.7864, 0.8178] \\

Proposed graph
& 0.7269
& 0.8207
& 0.8371
& 0.7707
& 0.7853
& 0.0000
& 0.7487 [0.7321, 0.7647] \\

\textbf{XGBoost pairwise}
& \textbf{0.8843}
& \textbf{0.9269}
& \textbf{0.9363}
& \textbf{0.9048}
& \textbf{0.9114}
& 0.0000
& \textbf{0.8747 [0.8632, 0.8861]} \\

\bottomrule
\end{tabular*}
\end{table*}

\begin{figure}[pos=tbp]
    \centering
    \includegraphics[
        width=0.96\columnwidth,
        keepaspectratio
    ]{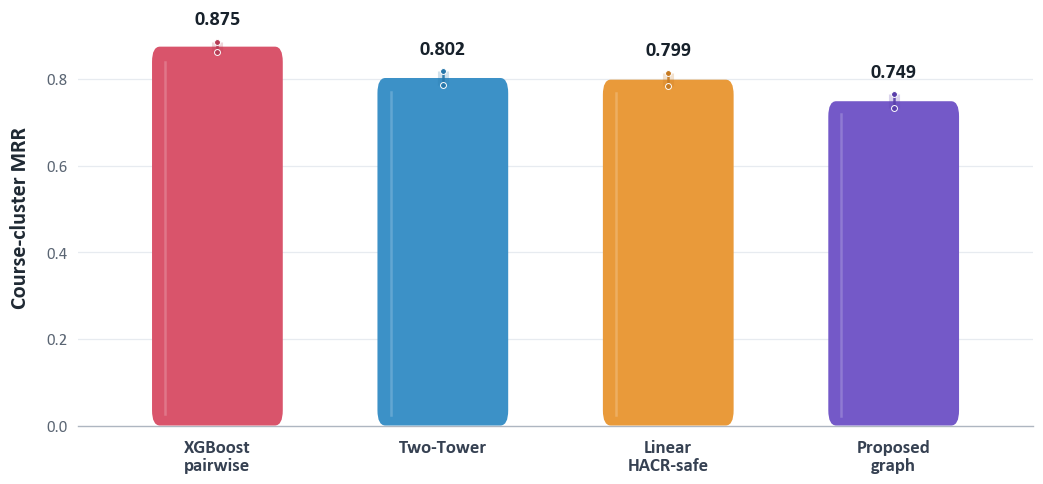}
    \caption{Frozen confirmatory course-cluster MRR with 95\% confidence intervals across 541 held-out MOOCCubeX courses. Course, rather than transition, is the resampling unit.}
    \label{fig:confirmatory_course_ci}
\end{figure}

Course-cluster aggregation produced the same principal ordering while treating course, rather than individual transition, as the inferential unit. XGBoost pairwise had a course-cluster mean MRR of 0.8747 (95\% CI: 0.8632--0.8861), compared with 0.8023 (0.7864--0.8178) for Two-Tower, 0.7985 (0.7831--0.8140) for Linear HACR-safe and 0.7487 (0.7321--0.7647) for the proposed graph model. Figure~\ref{fig:confirmatory_course_ci} displays these course-level estimates and intervals.

The paired course-level difference between the proposed graph model and XGBoost pairwise was \(-0.1260\) MRR across 541 courses, with 19 wins, 502 losses and 20 ties for the graph model. The two-sided Wilcoxon signed-rank test gave \(p=3.09\times10^{-81}\) before Holm adjustment and \(p=9.26\times10^{-81}\) after adjustment. Against Two-Tower, the corresponding mean difference was \(-0.0536\), with 87 wins, 436 losses and 18 ties (\(p=1.52\times10^{-42}\)). These inferential comparisons concern course-level ranking differences and do not alter the frozen model-selection procedure.

\subsection{Graph, Constraint, Feature, and Robustness Analyses}

The controlled development analyses separated the effects of graph inclusion, constraint semantics, allowed feature classes and graph perturbation. In the equal-budget graph-variant ablation, immediate next-resource MRR was highest without graph propagation: 0.8552 for the no-graph variant, compared with 0.8462 for the original graph, 0.8337 for the validated graph and 0.8230 for the refined graph. Thus, progressively processed prerequisite structure did not improve immediate MRR relative to the no-graph variant in this development comparison.

Constraint-policy behaviour depended on whether only inference semantics were altered or the model was retrained under each policy. With the relevance model frozen, hard-high-confidence produced MRR 0.7702, followed by soft at 0.7694, strict-original at 0.7683 and hybrid at 0.7668. Infeasible-target rates were 0.0056, 0, 0.0093 and 0.0056, respectively. Under equal-budget policy-specific retraining, hard-high-confidence yielded MRR 0.8193, hybrid 0.8184, soft 0.7952 and strict-original 0.7665. These are distinct experiments and are therefore reported separately in Table~\ref{tab:structural_analyses}.

\begin{table*}[tbp]
\centering
\caption{Development ablations of prerequisite structure, constraint treatment, and leakage-safe feature reliance.}
\label{tab:structural_analyses}

\footnotesize
\setlength{\tabcolsep}{7pt}
\renewcommand{\arraystretch}{1.10}

\begin{tabularx}{\textwidth}{
>{\raggedright\arraybackslash}p{2.5cm}
>{\raggedright\arraybackslash}X
>{\centering\arraybackslash}p{1.6cm}
>{\centering\arraybackslash}p{2.6cm}}
\toprule
\textbf{Model treatment} &
\textbf{Condition} &
\textbf{MRR} &
\textbf{Infeasible-target rate} \\
\midrule

\multicolumn{4}{l}{
\textit{A. Prerequisite-graph variants --- equal-budget development ablation}
} \\
\addlinespace[1pt]
-- & No graph  & 0.8552 & 0.0000 \\
-- & Original  & 0.8462 & 0.0096 \\
-- & Validated & 0.8337 & 0.0046 \\
-- & Refined   & 0.8230 & 0.0046 \\

\addlinespace[2pt]
\midrule
\multicolumn{4}{l}{
\textit{B. Constraint policies --- frozen inference and equal-budget retraining}
} \\
\addlinespace[1pt]
Frozen &
Strict-original &
0.7683 & 0.0093 \\
&
Soft &
0.7694 & 0.0000 \\
&
Hard-high-confidence &
0.7702 & 0.0056 \\
&
Hybrid &
0.7668 & 0.0056 \\
\addlinespace[1pt]
Retrained &
Strict-original &
0.7665 & 0.0096 \\
&
Soft &
0.7952 & 0.0000 \\
&
Hard-high-confidence &
0.8193 & 0.0046 \\
&
Hybrid &
0.8184 & 0.0046 \\

\addlinespace[2pt]
\midrule
\multicolumn{4}{l}{
\textit{C. Leakage-safe feature reliance --- frozen feature audit}
} \\
\addlinespace[1pt]
-- & All safe features                   & 0.7668 & 0.0056 \\
-- & Remove title features               & 0.0076 & 0.0056 \\
-- & Remove popularity                   & 0.7679 & 0.0056 \\
-- & Remove depth                        & 0.7891 & 0.0056 \\
-- & Remove prerequisite static features & 0.7702 & 0.0056 \\
-- & Concept/graph only                  & 0.0091 & 0.0056 \\

\bottomrule
\end{tabularx}

\vspace{2pt}
\begin{minipage}{\textwidth}
\footnotesize
\textit{Note.}
Viol@1 was 0 for all evaluated conditions and is therefore omitted.
Panel B evaluates constraint-policy changes under frozen inference and
equal-budget retraining. In Panel C, ``Concept/graph only'' masks all
16 leakage-safe static features.
\end{minipage}
\end{table*}

The feature-reliance audit showed particularly strong dependence on retained title/path-similarity information for immediate prediction. The all-safe setting produced MRR 0.7668, whereas masking the two title-derived safe features reduced MRR to 0.0076. By contrast, removing popularity produced 0.7679, removing depth features 0.7891 and removing prerequisite-static features 0.7702. The concept/graph-only condition produced MRR 0.0091. These observations concern permitted feature classes; direct candidate-order, order-distance, same-course and target-derived fields were excluded by the leakage-control policy.

\begin{figure}[pos=tbp]
    \centering
    \includegraphics[
        width=0.96\columnwidth,
        keepaspectratio
    ]{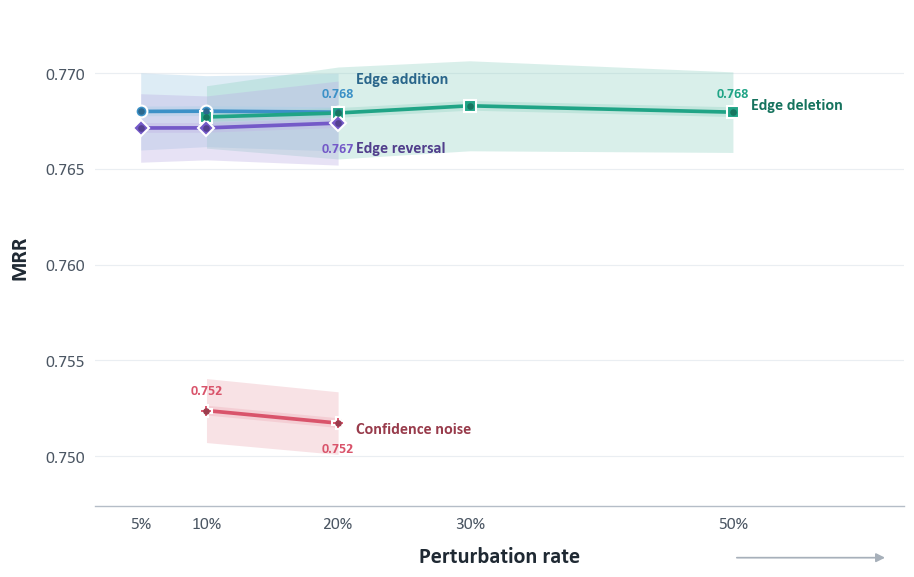}
    \caption{Development graph-robustness analysis across edge addition, deletion, reversal, and confidence-noise perturbations. Points summarize the prespecified perturbation conditions across three graph seeds; the complete condition grid is provided in the supplementary material.}
    \label{fig:graph_robustness}
\end{figure}

Graph perturbations produced different responses by corruption type. Across three seeds, added-edge conditions yielded mean MRRs of 0.7680 at perturbation rates 0.05, 0.10 and 0.20, while deletion conditions ranged from 0.7677 to 0.7683 over rates 0.10--0.50. Edge reversal produced MRRs between 0.7671 and 0.7674 over rates 0.05--0.20. Confidence perturbation produced the larger observed change: MRR decreased to 0.7524 at noise scale 0.10 and 0.7517 at 0.20, while mean infeasible-target rate increased to 0.0387 and 0.0401, respectively. Figure~\ref{fig:graph_robustness} presents the complete perturbation trends retained for the main text.

\subsection{Autoregressive Learning-Path Evaluation}

The frozen path protocol selected 60 courses from the frozen confirmatory set, but completed path artifacts contained 40 model/method rows spanning 12 distinct courses. All 60 selected courses were represented in the path-status record, while only 12 had at least one completed generated-path row. Path results are therefore reported using realised denominators rather than being described as evaluation on 60 completed courses.

Table~\ref{tab:confirmatory_paths} summarises the four realised model/rollout combinations. Concept-continuity greedy had the highest mean goal coverage among these completed rows at 0.5999 across 12 courses, followed by Random-valid greedy at 0.5254. Proposed-graph beam yielded mean goal coverage 0.4004 across four realised courses, whereas proposed-graph greedy yielded 0.2838 across 12. Mean path length was 45.83 for the three greedy summaries and 50.00 for the four realised beam paths. Termination success was 0.0833 for each greedy summary and 0 for the realised beam summary.

\begin{table*}[pos=tbp]
\centering
\caption{Realised autoregressive path performance on the frozen confirmatory set.}
\label{tab:confirmatory_paths}
\scriptsize
\setlength{\tabcolsep}{3.0pt}
\renewcommand{\arraystretch}{1.08}
\begin{tabularx}{\textwidth}{
>{\raggedright\arraybackslash}p{2.2cm}
>{\centering\arraybackslash}p{1.2cm}
>{\centering\arraybackslash}p{1.25cm}
>{\centering\arraybackslash}p{1.45cm}
>{\centering\arraybackslash}p{1.45cm}
>{\centering\arraybackslash}p{1.55cm}
>{\centering\arraybackslash}p{1.35cm}
>{\centering\arraybackslash}p{1.55cm}
>{\centering\arraybackslash}X}
\toprule
Model & Rollout & Realised courses & Goal coverage & Termination success & Prereq.\ violation & Redundancy & Concept diversity / step & Path efficiency \\
\midrule
Concept-continuity & Greedy & 12 & 0.5999 & 0.0833 & 0.0000 & 0.0596 & 20.2900 & 2.2367 \\
Random-valid       & Greedy & 12 & 0.5254 & 0.0833 & 0.0000 & 0.0157 & 5.6750  & 1.5617 \\
Proposed graph     & Greedy & 12 & 0.2838 & 0.0833 & 0.0000 & 0.0843 & 1.8933  & 0.5583 \\
Proposed graph     & Beam   & 4  & 0.4004 & 0.0000 & 0.0000 & 0.1240 & 2.9200  & 2.1850 \\
\bottomrule
\end{tabularx}

\vspace{2pt}
\parbox{\textwidth}{\footnotesize
The frozen path protocol selected 60 courses, but realised model/method coverage is reported exactly from the completed path outputs and should not be interpreted as complete coverage of all selected courses. The final path evidence contains 40 model/method rows across 12 distinct courses.}
\end{table*}

All four realised summary rows had mean prerequisite-violation rate 0.0000 and no repeated-resource accumulation under the frozen rollout policy, but other path properties differed. The proposed-graph beam summary had authored-suffix resource Jaccard 0.1051 across four realised courses, compared with 0.0175 for proposed-graph greedy, 0.0339 for concept-continuity and 0.0367 for Random-valid. Because the beam result contains only four completed paths, its denominator is reported explicitly rather than treated as equivalent to the 12-course greedy summaries.

\begin{figure*}[pos=tbp]
    \centering
    \includegraphics[
        width=0.94\textwidth,
        keepaspectratio
    ]{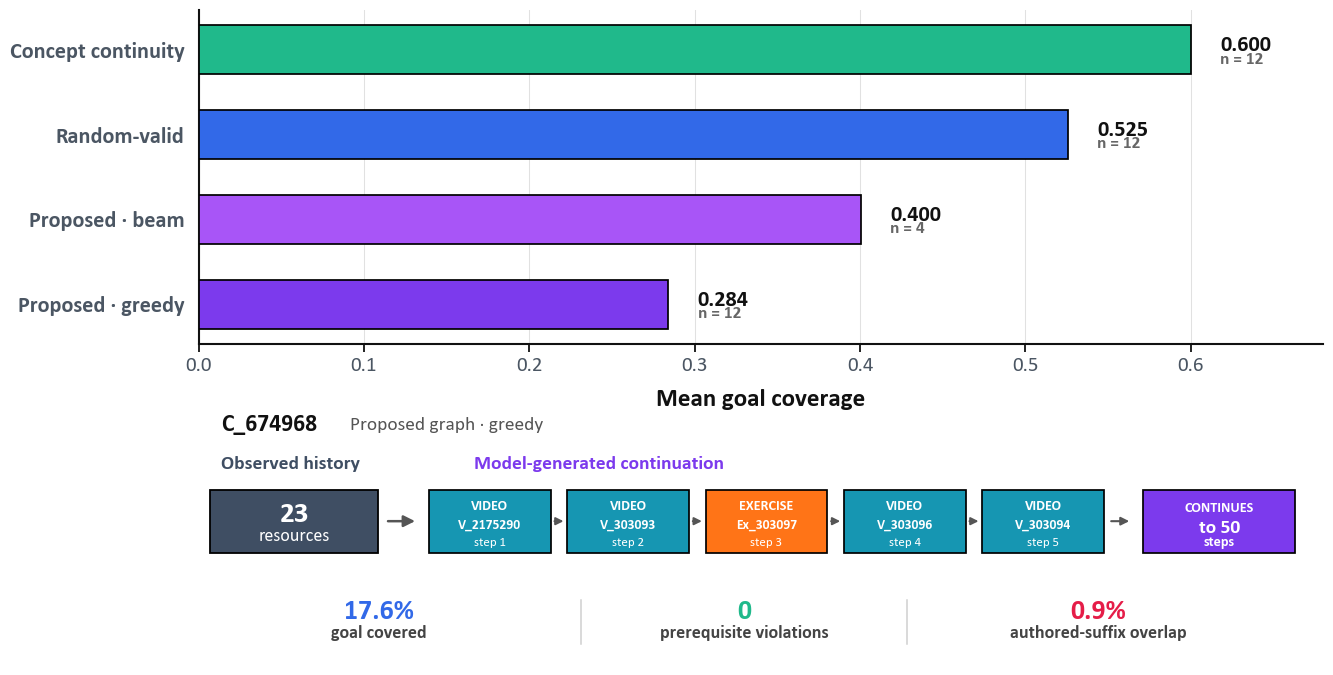}
    \caption{Long-horizon prerequisite-aware path behaviour: (a) aggregate realised confirmatory path goal coverage and (b) an illustrative confirmatory generated path for course \texttt{C\_674968} with structural outcome information. Panel (b) is a model-generated system output and was not human validated.}
    \label{fig:path_behaviour}
\end{figure*}

Figure~\ref{fig:path_behaviour} combines the aggregate confirmatory goal-coverage evidence with one existing model-generated output for course \texttt{C\_\allowbreak{}674968}. The illustrative panel uses the saved proposed-graph greedy rollout: prefix length 23, generated path length 50, goal coverage 0.1756, prerequisite-violation rate 0, termination success 0 and authored-suffix resource Jaccard 0.0086. Its first generated resources are \texttt{V\_\allowbreak{}2175290}, \texttt{V\_\allowbreak{}303093}, \texttt{Ex\_\allowbreak{}303097}, \texttt{V\_\allowbreak{}303096} and \texttt{V\_\allowbreak{}303094}. The example is included only as a saved system output and is not treated as representative or human validated.

\subsection{Junyi External Validation, Efficiency, and Explanation Evidence}

Junyi is reported as structure-/hierarchy-aware external validation; its Info\_Content hierarchy is tree metadata rather than an explicit prerequisite graph. Ranking used 250,000 held-out transitions representing 10,131 unique learners from the 10,151-learner test split and the complete 1,330-item catalog. Models were trained and selected within Junyi rather than transferred zero-shot from MOOCCubeX.

Markov achieved the highest observed Junyi MRR at 0.4453, with HR@1 0.2853 and NDCG@10 0.5118. Proposed-graph soft obtained MRR 0.3954, proposed graph 0.3954 and Two-Tower no-graph 0.3946. The proposed graph nevertheless had higher HR@10 than Markov, 0.7802 versus 0.7426. The paired learner-level difference in MRR between proposed graph and Markov was \(-0.0500\) (95\% CI: \(-0.0532,-0.0467\)), while the corresponding HR@10 difference was \(+0.0376\) (0.0345--0.0411). Thus, the external study again showed that explicit structural modelling did not automatically yield the highest immediate next-exercise ranking accuracy.

\begin{table*}[pos=tbp]
\centering
\caption{Junyi structure-/hierarchy-aware external validation.}
\label{tab:junyi_external}
\scriptsize
\setlength{\tabcolsep}{3.2pt}
\renewcommand{\arraystretch}{1.08}
\begin{tabularx}{\textwidth}{
>{\raggedright\arraybackslash}p{2.5cm}
>{\centering\arraybackslash}p{1.2cm}
>{\centering\arraybackslash}p{1.2cm}
>{\centering\arraybackslash}p{1.2cm}
>{\centering\arraybackslash}p{1.2cm}
>{\centering\arraybackslash}p{1.35cm}
>{\centering\arraybackslash}X}
\toprule
\multicolumn{7}{l}{\textit{Panel A: full-catalog ranking}}\\
Model & HR@1 & HR@5 & HR@10 & MRR & NDCG@10 & MRR 95\% CI \\
\midrule
Random                & 0.0007 & 0.0035 & 0.0075 & 0.0057 & 0.0033 & [0.0056, 0.0059] \\
Popularity            & 0.0262 & 0.0860 & 0.1294 & 0.0610 & 0.0706 & [0.0587, 0.0635] \\
Hierarchy continuity  & 0.0197 & 0.0981 & 0.1852 & 0.0760 & 0.0857 & [0.0743, 0.0776] \\
Two-Tower no-graph    & 0.2158 & 0.6336 & 0.7803 & 0.3946 & 0.4797 & [0.3912, 0.3980] \\
Proposed graph        & 0.2170 & 0.6355 & 0.7802 & 0.3954 & 0.4802 & [0.3923, 0.3988] \\
Proposed graph soft   & 0.2173 & 0.6348 & 0.7806 & 0.3954 & 0.4804 & [0.3923, 0.3987] \\
Markov                & 0.2853 & 0.6498 & 0.7426 & 0.4453 & 0.5118 & [0.4414, 0.4492] \\
\midrule
\multicolumn{7}{l}{\textit{Panel B: external path behaviour, \(n=1{,}000\) learners per model}}\\
Model & \multicolumn{2}{c}{Exact-step match} & \multicolumn{2}{c}{Suffix recall} & Hierarchy coherence & Branch diversity \\
\midrule
Popularity            & \multicolumn{2}{c}{0.0152} & \multicolumn{2}{c}{0.1324} & 0.7000 & 0.1009 \\
Hierarchy continuity  & \multicolumn{2}{c}{0.0185} & \multicolumn{2}{c}{0.1506} & 0.6979 & 0.1009 \\
Two-Tower no-graph    & \multicolumn{2}{c}{0.0787} & \multicolumn{2}{c}{0.5693} & 0.6877 & 0.1113 \\
Proposed graph        & \multicolumn{2}{c}{0.0806} & \multicolumn{2}{c}{0.5732} & 0.6879 & 0.1104 \\
Proposed graph soft   & \multicolumn{2}{c}{0.0811} & \multicolumn{2}{c}{0.5756} & 0.6892 & 0.1084 \\
Markov                & \multicolumn{2}{c}{0.0962} & \multicolumn{2}{c}{0.5907} & 0.6826 & 0.1147 \\
\bottomrule
\end{tabularx}
\end{table*}

The 1,000-learner external path analysis likewise produced metric-dependent differences. Markov had the highest exact-step match (0.0962) and suffix recall (0.5907), followed by proposed-graph soft at 0.0811 and 0.5756, and proposed graph at 0.0806 and 0.5732. Hierarchy coherence was 0.7000 for Popularity, 0.6979 for Hierarchy continuity, 0.6892 for proposed-graph soft, 0.6879 for proposed graph and 0.6826 for Markov. All retained path models had a unique-resource ratio of 1.0.

Efficiency measurements provide secondary computational context. The main MOOCCubeX graph-model training run required 6,298.6\,s, contained 15,911,634 parameters, reached 9.77\,GB peak CPU RSS, and recorded approximately 211.0\,Wh of GPU energy under the specified measurement procedure. The Two-Tower training run required 3,721.9\,s, contained 13,618,369 parameters and recorded approximately 124.0\,Wh. Across repeated development CV, median wall time was 1,697.1\,s for \texttt{graph\_\allowbreak{}no\_\allowbreak{}curriculum} and 158.3\,s for XGBoost pairwise. The Junyi graph model contained 246,451 parameters; its throughput varied with evaluation batch size. Finally, system-generated recommendation-explanation artifacts contained 1,150 development rows and 750 confirmatory-set rows. These outputs were not human-, instructor- or trust-validated.

\begin{figure}[pos=tbp]
    \centering
    \includegraphics[
        width=0.96\columnwidth,
        keepaspectratio
    ]{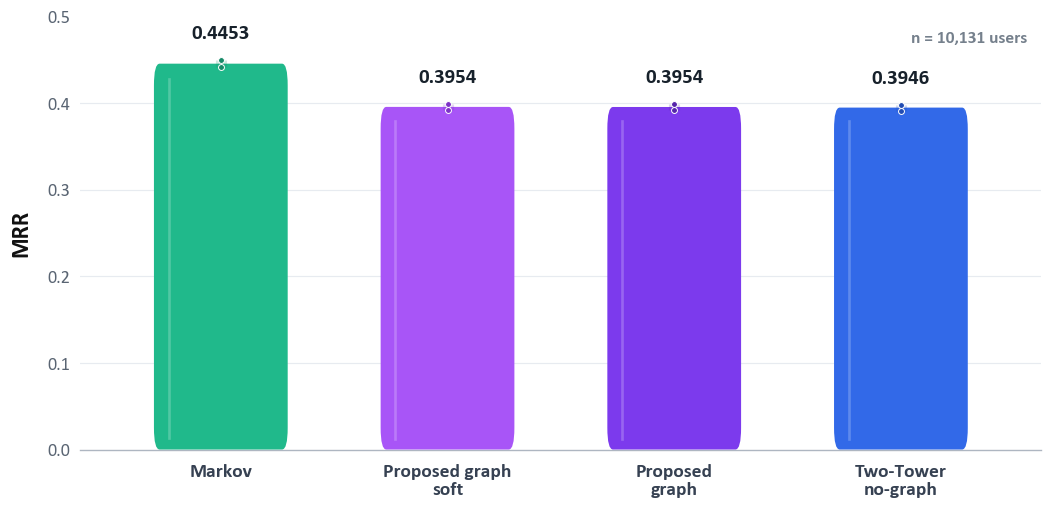}
\caption{Junyi structure-/hierarchy-aware external-validation MRR with learner-level 95\% confidence intervals. The evaluation uses independently trained Junyi models and does not constitute zero-shot transfer of MOOCCubeX model weights or prerequisite structure.}
\label{fig:junyi_external_ci}
\end{figure}

Across the complete evaluation, development-only course-level validation selected the no-curriculum graph configuration before the confirmatory holdout was scored. On the frozen confirmatory set, XGBoost pairwise provided the strongest observed immediate ranking performance, while the proposed graph model remained below XGBoost, Two-Tower and Linear HACR-safe on MRR. Development ablations likewise showed that adding prerequisite structure or changing constraint semantics did not uniformly increase immediate ranking accuracy, and the safe-feature audit showed strong dependence on permitted title/path-similarity signals. Autoregressive evaluation separately quantified long-horizon structural behaviour over 40 realised model/method rows spanning 12 courses from the 60-course protocol selection. Junyi external validation again favoured Markov on MRR, while structural and path metrics exhibited distinct model orderings.

\section{Discussion}

The findings support a reliability-oriented interpretation of prerequisite-aware educational recommendation in which immediate predictive accuracy, structural consistency and long-horizon path behaviour are treated as related but non-equivalent evaluation dimensions. The evidence also defines clear boundaries on what can be concluded from the present offline study.

\subsection{Principal Findings}

Four findings are central. First, prerequisite-aware graph modelling was not the strongest approach for immediate next-resource prediction. XGBoost pairwise provided the strongest observed confirmatory ranking performance, while the Two-Tower and linear baselines also exceeded the proposed graph model on immediate MRR. This result indicates that authored next-resource prediction contains substantial local, tabular and sequential signal that can be captured effectively without explicit graph propagation. It therefore limits any claim that prerequisite reasoning is necessary for maximising one-step ranking accuracy under the present protocol.

Second, structural modelling affected recommendation behaviour without uniformly improving immediate ranking. The no-graph development ablation obtained the strongest immediate MRR among the graph-variant conditions, while alternative hard, soft and hybrid constraint treatments produced different ranking and infeasible-target behaviours. Perturbation experiments further showed sensitivity to changes in structural information, particularly confidence perturbation. These results suggest that prerequisite structure should be treated as an uncertain modelling input whose consequences require explicit evaluation rather than as an automatically beneficial source of predictive information.

Third, the safe-feature analysis revealed pronounced dependence on retained title/path-similarity information. Removing the permitted title-derived features substantially reduced immediate ranking performance. Because direct course identity, candidate order, order distance and target-derived shortcuts were explicitly excluded, this finding should not be described as demonstrated leakage. Instead, it indicates strong proxy sensitivity to allowable local curricular signals, which may encode substantial information about authored resource organisation.

Fourth, autoregressive evaluation changes the question being asked. One-step ranking rewards agreement with the authored next resource, while path evaluation examines prerequisite consistency, goal coverage, redundancy, diversity, progression and termination. A method can therefore reproduce authored continuation effectively without necessarily providing the strongest representation of long-horizon structural behaviour, while structural reasoning cannot be assumed superior merely because it addresses pedagogically motivated constraints. Immediate predictive ranking and prerequisite-aware long-horizon reliability should consequently be evaluated as complementary objectives.

\subsection{Interpretation in Relation to Prior Work}

The findings can first be interpreted against the established distinction between sequence prediction and pedagogically structured recommendation. Previous educational recommenders have shown that observed interaction or course sequences contain substantial information for predicting subsequent resources. The strong XGBoost, Two-Tower and Junyi Markov results are consistent with that broader pattern. When the supervised target is the observed or authored next item, exploiting local sequence regularities is a legitimate and effective modelling strategy rather than a methodological deficiency.

A second contrast concerns complete-path and constraint-aware recommendation. Prior learning-path research has treated sequencing as a broader optimisation problem involving dependencies, path constraints, learner or scenario requirements and longer-term objectives. The present results reinforce the importance of maintaining this distinction. Immediate MRR describes agreement with the next authored resource, but it cannot establish whether recursively generated recommendations preserve structural feasibility, avoid redundancy, expand concept coverage or terminate appropriately. The value of path-level evaluation therefore lies in measuring properties that are outside the scope of conventional one-step ranking metrics, rather than in providing an alternative route to claiming predictive superiority.

The third contrast concerns the reliability of the knowledge structure itself. Knowledge graphs and prerequisite relations are often incorporated as modelling inputs, yet observed transitions, semantic associations and pedagogically valid prerequisites are not interchangeable. The no-graph ablation, constraint-policy differences and controlled perturbations are consistent with treating structural knowledge as uncertain rather than unquestioned ground truth. They do not establish that the prerequisite relations are pedagogically invalid, nor do they establish that the automated refinement procedure recovered a definitive prerequisite structure. Instead, they show that structural assumptions can materially affect recommendation behaviour and should therefore be audited and stress-tested. The Junyi results support the same broader caution from a different setting: its hierarchy provides useful educational structure, but its semantics differ from the prerequisite relations used in MOOCCubeX.

\subsection{Implications for Reliable Educational AI}

Several methodological implications follow from these findings. Strong non-graph baselines are important because weak comparisons could incorrectly attribute predictive gains to structural reasoning when local or tabular information is already sufficient for the immediate task. The observed XGBoost and Markov results show why conventional and sequential alternatives should remain part of reliability-oriented evaluation.

Ranking accuracy should also be separated explicitly from path-quality assessment. HR, MRR and NDCG quantify predictive agreement, whereas prerequisite feasibility, coverage, redundancy, diversity, progression and termination describe different properties of generated sequences. Neither class of measures subsumes the other.

Where educational recommenders depend on knowledge structures, controlled stress tests can provide additional evidence about reliability. Alternative constraint semantics and missing, added, reversed or confidence-perturbed relations help reveal whether conclusions depend strongly on a particular structural assumption.

Evaluation governance is similarly relevant. Course-level separation, repeated development validation, frozen confirmatory testing and explicit recovery provenance reduce opportunities for post-hoc model selection and make adverse findings easier to retain. In this study, the strong conventional baselines, no-graph result and external Markov result are informative precisely because the evaluation design does not require the proposed graph model to dominate every metric. Such practices support more precise claims about where structural modelling contributes and where its value remains uncertain.

\subsection{Limitations}

Several limitations bound the interpretation. The primary MOOCCubeX task is structural and course-prefix-conditioned rather than fully learner-personalised; it does not model individual mastery, preferences or response histories. Concept mapping is incomplete, and prerequisite relations cover only part of the evaluation space, so graph-based conclusions should not be generalised uniformly to every transition.

Long-horizon evidence is also incomplete relative to the planned protocol. Although 60 courses from the frozen confirmatory set were selected for path evaluation, completed artifacts contain 40 model/method rows across only 12 distinct courses. Path-level findings must therefore be interpreted within this realised coverage rather than as evidence over all selected courses.

The strong dependence on retained title/path-similarity signals further limits interpretation of immediate ranking as pure prerequisite reasoning, although it does not by itself establish leakage. External validation also differs structurally from the primary setting: Junyi uses hierarchy/tree metadata rather than an explicit prerequisite graph, and its models were trained within Junyi rather than transferred zero-shot. Finally, the study contains no completed expert validation of prerequisite edges, instructor-trust evaluation, learner intervention or measured learning-gain experiment. Offline structural and ranking metrics therefore do not establish pedagogical effectiveness or improved learner outcomes.

\subsection{Future Work}

Future work should address these limitations directly. Prerequisite structures could be evaluated through domain-expert review, explicit direction validation, confidence calibration and broader concept and edge coverage, providing a stronger basis for separating structural uncertainty from true pedagogical dependency.

The observed proxy sensitivity motivates experiments that reduce dependence on local title and sequence markers, including title normalisation, removal of numbering or hierarchical path tokens, stronger semantic representations and dedicated proxy-sensitivity evaluations. Handling of unseen or weakly represented concepts could also be extended through direct semantic embeddings, compositional concept encoders or graph components that are less dependent on a fixed development concept index.

A further direction is to combine stronger predictive relevance backbones with prerequisite reasoning as a gated or residual structural component, rather than requiring graph propagation to carry both immediate relevance and structural objectives. Longer-term evaluation should increase realised path coverage and incorporate instructor assessment of path validity, human evaluation of explanation usefulness and learner studies measuring educational outcomes. Finally, broader external evaluation across additional platforms, domains, knowledge structures and learner-history settings could examine whether the observed distinction between predictive ranking and structural reliability persists, including cross-dataset or zero-shot transfer settings where such evaluation is methodologically appropriate.

\section{Conclusion}

Learning-path recommendation requires more than reproducing the next authored resource because educational sequences may also need to respect prerequisite structure and remain coherent over multiple recommendation steps. This study therefore evaluated a reliability-oriented prerequisite-aware recommendation framework that combines resource and concept representations, confidence-aware prerequisite structure, alternative hard, soft and hybrid constraint semantics, immediate next-resource ranking, autoregressive path generation, graph perturbation analysis, repeated course-level development validation, frozen confirmatory evaluation, and independent structure-/hierarchy-aware validation on Junyi. The primary MOOCCubeX task remained structural and course-prefix-conditioned rather than learner-personalised.

The empirical evidence does not support a claim that prerequisite-aware graph modelling maximises immediate ranking accuracy. XGBoost pairwise was the strongest observed next-resource ranker on the frozen confirmatory set, while the Two-Tower and linear baselines also exceeded the proposed graph model on immediate MRR. Development ablations further showed that adding prerequisite structure or changing constraint treatment did not uniformly improve one-step ranking, and the safe-feature audit revealed strong dependence on retained local title/path-similarity signals. Autoregressive evaluation, however, exposed structural properties not measured by one-step ranking, including prerequisite consistency, coverage, redundancy, diversity, progression and termination behaviour. External validation on Junyi using hierarchical educational metadata similarly showed that structure-aware modelling did not automatically dominate a strong sequential baseline.

The central conclusion is therefore that immediate predictive accuracy, prerequisite-aware structural reliability and long-horizon path quality are complementary rather than interchangeable dimensions of educational recommendation. Progress will require stronger prerequisite validation and coverage, reduced dependence on local sequence proxies, improved unseen-concept handling, broader external and long-horizon evaluation, and human and learner-outcome validation. The contribution of this study is a controlled characterisation of these distinct dimensions under uncertain knowledge structures rather than a universal ranking-superiority claim.

\section*{Acknowledgments}

The authors acknowledge American International University-Bangladesh (AIUB)
for its institutional support. The authors also acknowledge the Advanced
Machine Intelligence Research Lab (AMIR) for its research support and
facilities.

\section*{Conflicts of Interest}

The authors declare that they have no known competing financial interests or
personal relationships that could have appeared to influence the work reported
in this paper.

\section*{Data and Code Availability Statement}

All datasets used in this study are publicly available from their official
repositories.

The MOOCCubeX dataset used for the main development and confirmatory
experiments is publicly available through the official repository maintained
by the Knowledge Engineering Group (THU-KEG), Tsinghua University:
\url{https://github.com/THU-KEG/MOOCCubeX}.

The Junyi Academy Online Learning Activity Dataset used for independent
cross-dataset validation is publicly available through the official Junyi
Academy dataset release:
\url{https://www.kaggle.com/datasets/junyiacademy/learning-activity-public-dataset-by-junyi-academy}.

No privately collected participant data were introduced in this study.
The course-level development and held-out split definitions, prerequisite
structure construction and perturbation procedures, Junyi learner-level split
protocol, preprocessing procedures, model configurations, training protocols,
and evaluation methodology used for the reported experiments are described
in the manuscript.

Code and processed experimental artefacts associated with this study will be
made publicly available through the project repository upon publication.

\section*{CRediT authorship contribution statement}

\noindent\textbf{M.\,M. Golam Hafiz:}
Conceptualization, Methodology, Software, Data Curation, Investigation,
Formal Analysis, Validation, Visualization, Writing -- Original Draft,
Writing -- Review \& Editing.

\noindent\textbf{A.K.M. Shahriar Emtiaz:}
Methodology, Software, Investigation, Visualization,
Writing -- Original Draft, Writing -- Review \& Editing.

\noindent\textbf{Safia Akter Simi:}
Investigation, Visualization, Writing -- Original Draft,
Writing -- Review \& Editing.

\noindent\textbf{Md. Rakibul Islam:}
Supervision, Project Administration, Methodology, Resources, Funding Acquisition,
Writing -- Review \& Editing.

\noindent\textbf{M.\,F. Mridha:}
Supervision, Project Administration, Methodology, Resources,
Writing -- Review \& Editing.

\section*{Declaration of Generative AI and AI-Assisted Technologies}

All experimental results, numerical values, statistical analyses, tables,
and figures reported in this study were produced directly from the authors'
implemented code, executed experiments, and verified experimental outputs.
No generative AI system was used to generate, modify, fabricate, or infer
experimental data, numerical results, statistical outcomes, plots, figures,
or scientific conclusions. No AI-based image-generation system was used to
create the manuscript figures or the methodological architecture diagram.

Generative AI tools were used only during manuscript preparation for
language refinement and organisational assistance. The authors independently
verified, revised, and take full responsibility for the final scientific
content of the manuscript.

\bibliographystyle{cas-model2-names}
\bibliography{CAEAI_References}

\end{document}